\pdfoutput=1

\PassOptionsToPackage{table}{xcolor} 
\documentclass[11pt]{article}

\usepackage[final]{acl}

\usepackage{times}
\usepackage{latexsym}

\usepackage[T1]{fontenc}

\usepackage[utf8]{inputenc}

\usepackage{microtype}

\usepackage{inconsolata}

\usepackage{graphicx}

\usepackage{amsmath}
\usepackage{amssymb}
\usepackage{bbm}
\usepackage{booktabs} 
\usepackage{fontawesome} 
\usepackage{makecell} 

\title{CoMoL: Efficient Mixture of LoRA Experts \\ via Dynamic Core Space Merging}

\author{
 \textbf{Jie Cao\textsuperscript{1}\Thanks{Equal contribution}},
 \textbf{Zhenxuan Fan\textsuperscript{1}\footnotemark[1]},
 \textbf{Zhuonan Wang\textsuperscript{1}},
 \textbf{Tianwei Lin\textsuperscript{1}},
 \textbf{Ziyuan Zhao\textsuperscript{2}},
 \\
 \textbf{Rolan Yan\textsuperscript{2}},
 \textbf{Wenqiao Zhang\textsuperscript{1}\Thanks{Corresponding author}},
 \textbf{Feifei Shao\textsuperscript{1}\footnotemark[2]},
 \textbf{Hongwei Wang\textsuperscript{1}},
 \textbf{Jun Xiao\textsuperscript{1}},
 \textbf{Siliang Tang\textsuperscript{1}}
\\
\\
 \textsuperscript{1}Zhejiang University
 \textsuperscript{2}Wechat, Tencent
\\
 \small{
   \{caojie, zxfan, wnzz, lintw, wenqiaozhang, sff, siliang\}@zju.edu.cn
 }
 \\
 \small{
 \{joshuazhao, rolanyan\}@tencent.com
 }, hongweiwang@intl.zju.edu.cn, junx@cs.zju.edu.cn 
}

\begin{document}
\maketitle

\begin{abstract}

Large language models (LLMs) achieve remarkable performance on diverse downstream and domain-specific tasks via parameter-efficient fine-tuning (PEFT). However, existing PEFT methods, particularly MoE-LoRA architectures, suffer from limited parameter efficiency and coarse-grained adaptation due to the proliferation of LoRA experts and instance-level routing. To address these issues, we propose Core Space Mixture of LoRA (\textbf{CoMoL}), a novel MoE-LoRA framework that incorporates expert diversity, parameter efficiency, and fine-grained adaptation. Specifically, CoMoL introduces two key components: Core Space Experts and Core Space Routing. Core Space Experts store each expert in a compact core matrix, preserving diversity while controlling parameter growth. Core Space Routing dynamically selects and activates the appropriate core experts for each token, enabling fine-grained, input-adaptive routing. Activated core experts are then merged via a soft-merging strategy into a single core expert, which is combined with a shared LoRA to form a specialized LoRA module. Besides, the routing network is projected into the same low-rank space as the LoRA matrices, further reducing parameter overhead without compromising expressiveness. Extensive experiments demonstrate that CoMoL preserves the adaptability of MoE-LoRA architectures while achieving parameter efficiency comparable to Standard LoRA, consistently outperforming existing methods across multiple tasks. 
Our code is available at 
\url{https://github.com/DCDmllm/CoMoL}.

\end{abstract}
\section{Introduction}
The rapid advancement of large language models (LLMs)~\citep{achiam2023gpt,yang2025qwen3,grattafiori2024llama} has greatly enhanced natural language processing across a variety of tasks. However, their massive size makes it challenging to adapt them efficiently to downstream or domain-specific applications. 
To address this, 
parameter-efficient fine-tuning (PEFT) methods \citep{houlsby_adapter_2019, he_towards_2022_paralleladapter, li_prefix-tuning_2021, lester_power_2021_prompttuning, hu_lora_2021} have emerged, enabling adaptation by updating a small fraction of model parameters, which substantially reduces training costs and alleviates catastrophic forgetting.
Among them, Low-Rank Adaptation (LoRA) \citep{hu_lora_2021} achieves competitive performance through low-rank decomposition of weight matrices with minimal trainable parameters. Unfortunately, recent studies reveal that naively increasing the LoRA rank does not lead to proportional performance gains \citep{chen2022revisiting, zhu_sira_2023}.


\begin{figure*}[ht]
  \includegraphics[width=1\textwidth]{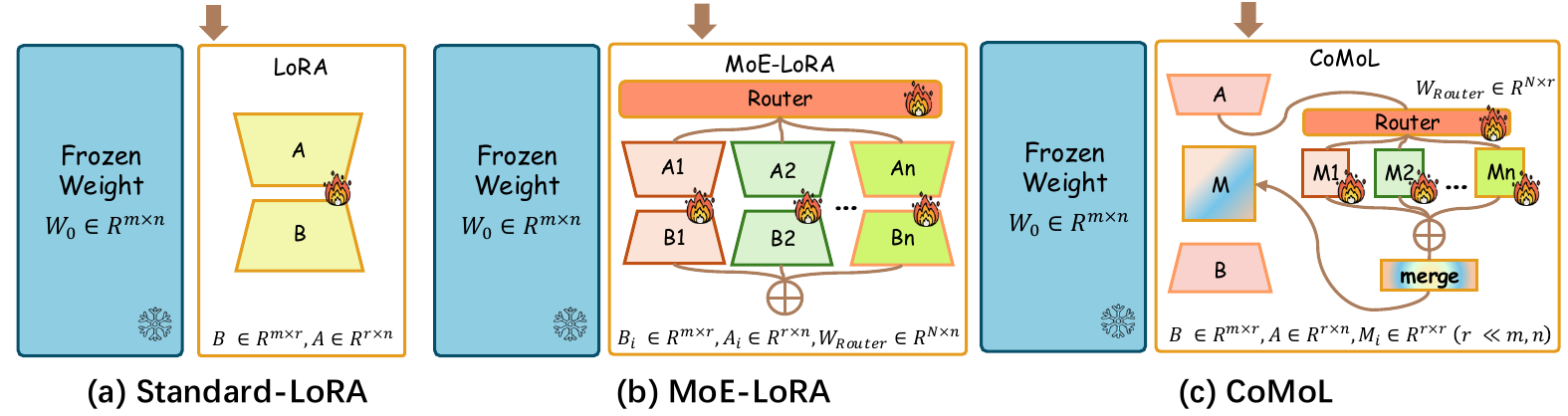}
  \caption{
    \textbf{Comparison of LoRA-based methods.} \textbf{(a)} LoRA injects trainable low-rank matrices $\boldsymbol{A}$ and $\boldsymbol{B}$ into transformer layers to approximate weight updates. \textbf{(b)} MoE-LoRA employs multiple LoRA experts with a routing network to select and activate experts based on input tokens. \textbf{(c)} CoMoL fuses expert parameters in the low-rank core space, preserving token-level routing while maintaining expert-specific expressiveness with minimal parameters.}
  \label{fig:CoMoL}
\end{figure*}

Inspired by Mixture-of-Experts (MoE)~\citep{shazeer2017outrageously_sparsemoe}, recent PEFT studies integrate LoRA into the MoE architecture to enhance model capacity, giving rise to the MoE-LoRA paradigm. Specifically, MoE-LoRA consists of a routing network and multiple LoRA experts, where the router dynamically selects and activates experts conditioned on the input condition \citep{zadouri_pushing_2023, zhu_sira_2023}. This design is motivated by two key considerations: \textbf{(1)} multiple LoRA experts can capture complementary representations, providing greater expressiveness than a single LoRA module; and \textbf{(2)} the routing network facilitates flexible, input-dependent expert selection, enabling more efficient representation learning. 




Despite these advancements, existing MoE-LoRA methods still face two major limitations. \textbf{(1) Limited Parameter Efficiency:} The MoE-LoRA architecture adds a routing network and multiple LoRA modules, most of which participate in subsequent computations, leading to increased parameter counts and higher computational overhead; \textbf{(2) Coarse-grained Adaptation:} 
SMEAR \cite{muqeeth_soft_2024} constructs a single merged expert by computing its parameters as a weighted average of multiple experts within a routing block.
However, its routing mechanism operates at the instance level, relying on instance features rather than token-level features, which limits fine-grained adaptability.

To address the challenges of MoE-LoRA architectures, we propose Core Space Mixture of LoRA (\textbf{CoMoL}), a novel MoE-LoRA framework that incorporates expert diversity, parameter efficiency, and fine-grained adaptation.
As illustrated in Figure~\ref{fig:CoMoL}(c), CoMoL introduces two key components: Core Space Experts and Core Space Routing. Core Space Experts store each expert in a compact core matrix within a shared core space, whose dimensionality matches the LoRA rank, preserving diverse expert knowledge while controlling parameter overhead. Core Space Routing dynamically selects and activates the appropriate core experts for each input token, enabling fine-grained, token-level adaptation. Activated experts are then merged through a soft-merging strategy into a single core expert, which is combined with a LoRA module to form a specialized LoRA. 





Compared to previous MoE-LoRA methods, CoMoL preserves expert knowledge with only a minimal set of parameters in the core space. Importantly, by performing token-level routing and soft-merging entirely within the core space, CoMoL enables fine-grained adaptation with minimal overhead. To further reduce the parameter cost of the routing module, CoMoL leverages the inherent low-rank structure of LoRA, projecting the router into the same low-rank space. This design simultaneously maintains strong performance and reduces the overall parameter count to a level comparable to Standard LoRA. 
The contributions of this paper are summarized as follows:

\begin{itemize}

    \item We highlight that existing MoE-LoRA methods primarily suffer from limited parameter efficiency and coarse-grained adaptation, stemming from the proliferation of LoRA experts and instance-level routing over experts.


    \item We introduce Core Space Mixture of LoRA (\textbf{CoMoL}), an enhanced MoE-LoRA framework that stores expert parameters within a compact core space, enabling token-level routing while achieving parameter efficiency comparable to Standard LoRA.

    \item Extensive experiments against state-of-the-art MoE-LoRA methods demonstrate that CoMoL achieves superior scalability and robustness across a wide range of tasks and settings.
\end{itemize}

\section{Revisiting MoE-LoRA Methods}
\subsection{LoRA Basics}
LoRA \cite{hu_lora_2021} injects trainable low-rank matrices into transformer layers to approximate the weight updates. For a frozen pre-trained weight matrix $\boldsymbol{W} \in \mathbb{R}^{m \times n}$, LoRA represents its update with a low-rank decomposition: 
\begin{equation}
    \boldsymbol{W} + \Delta \boldsymbol{W} = \boldsymbol{W} + \boldsymbol{B}\boldsymbol{A} ,
\end{equation}  
where $\boldsymbol{B} \in \mathbb{R}^{m \times r}$, $\boldsymbol{A} \in \mathbb{R}^{r \times n}$ are tunable parameters, and the rank $r \ll min(m,n)$. For an input token $\boldsymbol{x}$ to the linear projection in the transformer layer, the computation of LoRA is as follows:
\begin{equation}
    \boldsymbol{h} = \boldsymbol{Wx} +  \boldsymbol{BAx}.
\end{equation}

\begingroup
\begin{table*}[t]
  \centering
  \begin{tabular}{lcccc}
    \Xhline{1.5pt}
    \rowcolor[HTML]{DAE0FB}
    \textbf{Methods} & Params ($\times$) & FLOPs ($\times$) & Routing Level & Routing Latency \\
    \hline
    \hline
    LoRA & $1.0\times$ & $1.0\times$ & - & -\\
    Soft-weighted MoE-LoRA & $N\times$ & $N\times$ & Token & Low\\
    Sparse MoE-LoRA & $N\times$ & $k\times$ & Token & High \\
    Soft-merging MoE-LoRA & $N\times$ & $1.0\times$ & Instance & Low\\
    \hline
    \rowcolor[HTML]{E9F3FE}
    CoMoL & $1.0\times$ & $1.0\times$ & Token & Low \\
    \Xhline{1.5pt}
  \end{tabular}
  \caption{\label{table:moe-lora_comparison}
  Comparison of LoRA and MoE-LoRA variants. Metrics for "Params" and "FLOPs" reflect only the LoRA experts; routing overhead is omitted for simplicity. $N\times$ indicates the ratio relative to the LoRA baseline.
  }
\end{table*}

\subsection{MoE-LoRA Methods}
The MoE-LoRA structure consists of $N$ different LoRA experts $\{\boldsymbol{E}_1, \boldsymbol{E}_2, \dots, \boldsymbol{E}_N\}$, where each LoRA expert $\boldsymbol{E}_i$ is represented by a pair of matrices $\{\boldsymbol{B}_{i} \in \mathbb{R}^{m \times r}, \boldsymbol{A}_i \in \mathbb{R}^{r \times n}\}$. The MoE-LoRA structure incorporates a gating network (Router) $\boldsymbol{G}(\boldsymbol{x}):\mathbb{R}^{n} \rightarrow \mathbb{R}^{N}$ that computes the weight $\boldsymbol{G}(\boldsymbol{x})_i$ for each expert $\boldsymbol{E}_i$ given an input token $\boldsymbol{x}$.
The forward process of MoE-LoRA is computed as:
\begin{equation}
    \boldsymbol{h} = \boldsymbol{Wx} + \sum_{i \in \mathcal{T}(\boldsymbol{x})}\boldsymbol{G}(\boldsymbol{x})_{i} \cdot \boldsymbol{B}_{i}\boldsymbol{A}_{i}\boldsymbol{x},
\end{equation}
where the output $\boldsymbol{h}$ combines the frozen linear projection output and the router-weighted combination of activated expert outputs, $\mathcal{T}(\boldsymbol{x})$ denotes the set of indices for activated experts determined by the routing mechanism.

MoE-LoRA employs two types of routing mechanisms: \textit{soft routing mechanism} and \textit{sparse routing mechanism}, which correspond to \textit{soft-weighted MoE-LoRA} and \textit{sparse MoE-LoRA}, respectively.
\paragraph{Soft routing mechanism.}
Given an input token $\boldsymbol{x}$, the soft routing mechanism typically employs a dense linear layer followed by a softmax function:
\begin{equation}
    \boldsymbol{G}(\boldsymbol{x}) = \textrm{softmax}(\boldsymbol{W}_{g}\boldsymbol{x}),
\end{equation}
where $\boldsymbol{W}_{g} \in \mathbb{R}^{N \times n}$ is the trainable projection. Soft-weighted MoE-LoRA with soft routing mechanism activates all LoRA experts and combines their outputs weighted by the routing scores $\boldsymbol{G}(\boldsymbol{x})$.
\paragraph{Sparse routing mechanism.} The sparse MoE-LoRA selectively activates a subset of experts, typically the top-k LoRA experts based on routing scores $\boldsymbol{G}(\boldsymbol{x})$, and aggregates only the outputs from the activated experts:
\begin{equation}
    \boldsymbol{G}(\boldsymbol{x}) = \textrm{top-k}(\textrm{softmax}(\boldsymbol{W}_{g}\boldsymbol{x})).
\end{equation}

\subsection{Soft-merging MoE-LoRA Method}
To reduce the computational costs of LoRA experts, SMEAR \cite{muqeeth_soft_2024} proposes an instance-level LoRA expert fusion method. SMEAR first makes a single routing choice for an entire input example:
\begin{equation}
    \boldsymbol{G}(\hat{\boldsymbol{x}})^{ins} = \textrm{softmax}(\boldsymbol{W}_{g}\hat{\boldsymbol{x}}),
\end{equation}
where $\hat{\boldsymbol{x}}$ denotes the average representation of the input example. Then, based on the routing scores, the LoRA experts are merged into a single expert:
\begin{equation}
    \boldsymbol{B}_{ins} = \sum_{i=1}^{N}\boldsymbol{B}_{i}\boldsymbol{G}(\hat{\boldsymbol{x}})^{ins}_{i},
    \boldsymbol{A}_{ins} = \sum_{i=1}^{N}\boldsymbol{A}_{i}\boldsymbol{G}(\hat{\boldsymbol{x}})^{ins}_{i},
\end{equation}
Finally, in the forward process of SMEAR, the merged expert is applied to compute each token in the sample:
\begin{equation}
    \boldsymbol{h} = \boldsymbol{Wx} + \boldsymbol{B}_{ins}\boldsymbol{A}_{ins}\boldsymbol{x},
\end{equation}
It is important to note that while in other MoE-LoRA methods each token in a sample is processed by different expert combinations, in SMEAR, all tokens in the sample use the same merged expert.
\subsection{Comparison of MoE-LoRA Methods}
Table~\ref{table:moe-lora_comparison} provides a comprehensive comparison between soft-weighted, sparse, and soft-merging MoE-LoRA architectures, evaluated across parameters, FLOPs, routing granularity, and routing latency. All three variants maintain an $N\times$ increase in LoRA expert parameters, which fundamentally diverges from the core principles of Parameter-Efficient Fine-Tuning (PEFT). Furthermore, previous research \cite{tian_hydralora_2024,lin2024teamlora,cao2025moa} indicates that these multiple experts often suffer from significant parameter redundancy. 

While soft-weighted MoE-LoRA activates all experts, resulting in $N\times \textrm{FLOPs}$, sparse MoE-LoRA attempts to mitigate this by activating only a subset (typically top-2). However, the empirical analysis \cite{cao2025moa} reveals an efficiency paradox: despite its lower theoretical FLOPs, sparse MoE-LoRA incurs significantly higher wall-clock latency than its soft-weighted counterpart. Under identical experimental settings (e.g., $N=8$), sparse routing increases training time by approximately $2\times$ and inference time by $1.5\times$ \cite{cao2025moa}, primarily due to the substantial computational overhead of the sparse routing mechanism. Conversely, while soft-merging MoE-LoRA reduces the burden to a single merged expert computation, it sacrifices token-level dynamism by degrading to an instance-level approach, thus limiting the model's fine-grained expressive capacity.

Overall, an ideal MoE-LoRA variant should achieve a favorable trade-off between expressivity and efficiency. Specifically, it is imperative to:
\textbf{(i)} Minimize the total parameter footprint while preserving the specialized characteristics of individual LoRA experts;
\textbf{(ii)} Reduce the computational overhead associated with the activated experts to enhance overall efficiency;
\textbf{(iii)} Sustain a fine-grained token-level routing mechanism while ensuring the routing latency remains negligible.

\section{Technical Details of CoMoL}
\subsection{Core Space of LoRA}
Let $\{\boldsymbol{B} \in \mathbb{R}^{m \times r}, \boldsymbol{A} \in \mathbb{R}^{r \times n}\}$ denote a LoRA expert within the MoE-LoRA framework. Each expert-specific weight update, $\Delta \boldsymbol{W}=\boldsymbol{B}\boldsymbol{A}$, can be re-parameterized via the reduced SVD of its constituent matrices:
\begin{equation}
    \begin{aligned}
    &\boldsymbol{B}=\boldsymbol{U}_B \boldsymbol{\Sigma}_B \boldsymbol{V}_B^\top, \boldsymbol{A}=\boldsymbol{U}_A \boldsymbol{\Sigma}_A \boldsymbol{V}_A^\top, \\
    &\Delta \boldsymbol{W}=\boldsymbol{U}_B \boldsymbol{\Sigma}_B \boldsymbol{V}_B^\top \boldsymbol{U}_A \boldsymbol{\Sigma}_A \boldsymbol{V}_A^\top,
    \end{aligned}
\end{equation}
where the dimensions are partitioned as $\boldsymbol{U}_B \in \mathbb{R}^{m \times r}$, $\boldsymbol{\Sigma}_B \in \mathbb{R}^{r \times r}$, $\boldsymbol{V}_B^\top \in \mathbb{R}^{r \times r}$, $\boldsymbol{U}_A \in \mathbb{R}^{r \times r}$, $\boldsymbol{\Sigma}_A \in \mathbb{R}^{r \times r}$, and $\boldsymbol{V}_A^\top \in \mathbb{R}^{r \times n}$.

Following the definition of the \textbf{Core Matrix} $\boldsymbol{M} \in \mathbb{R}^{r \times r}$ \cite{panarielloaccurate_corespace}:
\begin{equation}
    \boldsymbol{M} = \boldsymbol{\Sigma}_B \boldsymbol{V}_B^\top \boldsymbol{U}_A \boldsymbol{\Sigma}_A,
\end{equation}
the expert weight update can be compactly decoupled as:
\begin{equation}
    \Delta \boldsymbol{W} = \boldsymbol{U}_B \boldsymbol{M} \boldsymbol{V}_A^\top.
\end{equation}
In this formulation, the matrices $\boldsymbol{U}_B$ and $\boldsymbol{V}_A^\top$ span the singular subspaces that determine the feature orientations, while the core matrix $\boldsymbol{M}$ modulates the coupling strength and weight allocation across these latent dimensions.

\subsection{Core Space Mixture of LoRA} \label{sec:flops}
To address the parameter redundancy inherent in MoE-LoRA experts, we hypothesize that individual experts share a substantially similar latent subspace defined by the singular bases $\boldsymbol{U}_B$ and $\boldsymbol{V}_A^\top$. By confining expert-specific adaptations exclusively to the compact core matrix $\boldsymbol{M}_i$, we achieve a significant reduction in the parameter footprint. Under this formulation, the forward process for a given input $\boldsymbol{x}$ is defined as:
\begin{equation} \label{eq:forward}
\boldsymbol{h} = \boldsymbol{Wx} + \sum_{i=1}^{N}\boldsymbol{G}(\boldsymbol{x})_{i} \cdot (\boldsymbol{U}_B \boldsymbol{M}_i \boldsymbol{V}_A^\top) \boldsymbol{x},
\end{equation}
where $G(\boldsymbol{x})_i$ is a scalar that denotes the token-level routing weight for the $i$-th expert.

\paragraph{Parameter Efficiency}
Unlike standard MoE-LoRA, which replicates the full $\{ \boldsymbol{B}_i, \boldsymbol{A}_i \}$ pairs $N$ times, our approach constrains the per-expert parameters to the core matrix $\boldsymbol{M}_i$. This reduces the expert-related parameter complexity from $\mathcal{O}(N \cdot (m+n)r)$ to $\mathcal{O}((m+n)r + N \cdot r^2)$. Since $r \ll m, n$, our method maintains a near-constant parameter footprint as the number of experts $N$ scales, effectively satisfying the core principles of PEFT.

\paragraph{Token-level soft-merging of experts in Core Space}
Another key advantage of our formulation is the ability to perform core space expert merging at the token level without significant computational overhead. By exploiting \textbf{the distributive property of matrix multiplication}, the forward process described by Equation~\ref{eq:forward} can be equivalently expressed as
\begin{equation} \label{eq:forword_softmerge}
\boldsymbol{h} = \boldsymbol{Wx} + \boldsymbol{U}_B \left( \sum_{i=1}^{N} \boldsymbol{G}(\boldsymbol{x})_i \boldsymbol{M}_i \right) \boldsymbol{V}_A^\top \boldsymbol{x}.
\end{equation}
This allows the model to first aggregate the small $r \times r$ core matrices into a single merged matrix $\boldsymbol{M}_{merged}$ for each token. Consequently, the high-dimensional projections $\boldsymbol{U}_B$ and $\boldsymbol{V}_A$ are only computed once per token, reducing the total FLOPs to a level comparable to a single LoRA layer while preserving the fine-grained token-level dynamism of the MoE mechanism.

To provide a rigorous computational complexity analysis, we compare the two expert aggregation strategies. Consider a batch of $L$ tokens with a model dimension $n$, a LoRA rank $r$, and $N$ total experts. Under the output-level fusion strategy (as defined in Eq.\ref{eq:forward}), the FLOPs required for expert computation scale as $2 \times L \times r \times (2n + r) \times N$, while the subsequent aggregation of expert outputs incurs $L \times n \times (N-1)$ FLOPs. In contrast, our proposed soft-merging in core space (Eq.~\ref{eq:forword_softmerge}) streamlines this process. The FLOPs for expert computation are reduced to $2 \times L \times r \times (2n + r)$, and the fusion of core matrices requires only $L \times r^2 \times (N-1)$ FLOPs. Notably, our Core Space Merging reduces the primary expert computational overhead to a factor of $1/N$ compared to the output-level baseline. Furthermore, given that $r^2 \ll n$, the cost of aggregating core matrices is significantly lower than that of fusing high-dimensional expert outputs. This demonstrates that our method achieves superior efficiency in both primary transformation and expert orchestration.

\subsection{Core Space Routing}
In conventional MoE-LoRA architectures, the parameter count of the router $\boldsymbol{W}_{g} \in \mathbb{R}^{N \times n}$ scales linearly with the number of experts $N$. This dependency introduces a significant computational bottleneck and a substantial memory footprint, thereby undermining the inherent efficiency of the MoE-LoRA framework. To circumvent this limitation, we propose a Core Space Routing mechanism, which significantly reduces the parameter overhead while effectively leveraging the low-rank properties of LoRA. Specifically, we reuse the intermediate result of the input $\boldsymbol{x}$ in the low-rank projection space, denoted as $\hat{\boldsymbol{x}} = \boldsymbol{V}_A^\top \boldsymbol{x} \in \mathbb{R}^{r}$, as the input for the router. The layer output $\boldsymbol{h}$ is then computed as follows:
\begin{equation}
    \hat{\boldsymbol{x}} = \boldsymbol{V}_A^\top \boldsymbol{x}
\end{equation}
\begin{equation}
    \boldsymbol{h} = \boldsymbol{Wx} + \boldsymbol{U}_B \left( \sum_{i=1}^{N} \boldsymbol{G}(\hat{\boldsymbol{x}})_i \boldsymbol{M}_i \right) \hat{\boldsymbol{x}}.
\end{equation}
By performing routing within this latent core space, the parameter complexity of the router weight $\boldsymbol{W}_{g}$ is reduced from $\mathcal{O}(N \cdot n)$ to $\mathcal{O}(N \cdot r)$. Given that the rank $r$ is typically orders of magnitude smaller than the model dimension $n$ (e.g., $r=8$ vs. $n=4096$), this mechanism drastically curtails the storage requirements and computational cost associated with the router without sacrificing the fine-grained dynamism of token-level routing.

\section{Experiments}

In this section, we present a comprehensive empirical evaluation to validate the effectiveness of CoMoL across a diverse range of benchmarks. We first assess its mathematical reasoning capabilities using the Qwen3-8B and its larger counterpart, Qwen3-14B, to examine performance across different parameter scales. Subsequently, to verify the cross-architecture robustness of our method, we evaluate CoMoL on coding tasks using representative models from distinct families, specifically Qwen3-8B and Llama3.1-8B.

To facilitate a rigorous and comprehensive comparative analysis, we benchmark our proposed method against Standard LoRA and several state-of-the-art MoE-LoRA variants across different paradigms: \textbf{(i)} the vanilla Standard LoRA \citep{hu_lora_2021}; \textbf{(ii)} soft-weighted MoE-LoRA methods, including MoLoRA \cite{zadouri_pushing_2023} and HydraLoRA \cite{tian_hydralora_2024}; \textbf{(iii)} sparse MoE-LoRA frameworks, represented by MoLA \citep{gao_higher_2024}, AdaMoLE \cite{liu_adamole_2024} and SparseMoA \cite{cao2025moa}; and \textbf{(iv)} other advanced LoRA-based methods such as DenseLoRA \cite{mu2025denselora} and FlyLoRA \cite{zouflylora}.

\subsection{Mathematical reasoning}

\begingroup
\setlength{\tabcolsep}{4pt}
\begin{table*}[t]
\small
  \centering 
  \begin{tabular}{lcccccccc}
    \Xhline{1.5pt}
    \rowcolor[HTML]{DAE0FB}
    \textbf{Models} & \textbf{AddSub} & \textbf{AQuA} & \textbf{GSM8k} & \textbf{MultiArith} & \textbf{SingleEq} & \textbf{SVAMP} & \textbf{Average} & \textbf{Param} \\
    \hline
    \hline
    \rowcolor[HTML]{D3D3D3}
    LoRA & $93.67_{\pm 0.39}$ & $37.40_{\pm 2.58}$ & $81.80_{\pm 0.67}$ & $98.17_{\pm 0.86}$ & $96.85_{\pm 0.23}$ & $88.80_{\pm 0.53}$ & $82.78_{\pm 0.47}$ & 24.77M \\ 
    MoLoRA & $93.76_{\pm 1.30}$ & $42.26_{\pm 2.41}$ & $82.51_{\pm 2.35}$ & $98.11_{\pm 0.38}$ & $96.06_{\pm 1.04}$ & $90.40_{\pm 0.30}$ & $83.85_{\pm 1.17}$ &  107.35M \\
    HydraLoRA & $93.16_{\pm 1.27}$ & $39.11_{\pm 2.79}$ & $81.78_{\pm 1.48}$ & $98.11_{\pm 0.25}$ & $95.21_{\pm 0.75}$ & $88.80_{\pm 0.26}$ & $82.70_{\pm 0.88}$ & 49.55M\\
    MoLA & $93.33_{\pm 0.89}$ & $42.78_{\pm 0.45}$ & $81.50_{\pm 1.14}$ & $98.33_{\pm 0.44}$ & $96.13_{\pm 0.50}$ & $89.83_{\pm 1.16}$ & $83.65_{\pm 0.26}$ & 107.35M \\
    AdaMoLE & $92.24_{\pm 1.05}$ & $41.73_{\pm 4.26}$ & $81.65_{\pm 1.00}$ & $98.22_{\pm 0.25}$ & $96.19_{\pm 0.80}$ & $89.10_{\pm 0.75}$ & $83.19_{\pm 0.85}$ & 108.38M \\ 
    SparseMoA &  $93.33_{\pm 0.15}$ & $40.68_{\pm 2.95}$ & $83.85_{\pm 0.35}$ & $98.17_{\pm 0.44}$ & $97.31_{\pm 0.11}$ & $88.80_{\pm 1.66}$ & $83.69_{\pm 0.76}$ &  24.49M \\ 
    FlyLoRA & $93.00_{\pm 0.81}$ & $39.11_{\pm 1.98}$ & $86.20_{\pm 0.50}$ & $98.44_{\pm 0.10}$ & $95.87_{\pm 0.86}$ & $89.23_{\pm 0.70}$ & $83.64_{\pm 0.40}$ & 24.77M \\
    \hline
    \rowcolor[HTML]{E9F3FE}
    CoMoL w/o CR & $93.92_{\pm 2.42}$ & $44.09_{\pm 1.80}$ & $83.67_{\pm 2.51}$ & $98.28_{\pm 0.19}$ & $97.31_{\pm 0.69}$ & $89.53_{\pm 0.80}$ & $\underline{84.47}_{\pm 1.25}$ & 33.39M\\
    \rowcolor[HTML]{E9F3FE}
    CoMoL & $94.60_{\pm 1.05}$ & $44.62_{\pm 5.32}$ & $83.93_{\pm 1.64}$ & $98.22_{\pm 0.69}$ & $96.46_{\pm 0.86}$ & $89.03_{\pm 0.55}$ & $\boldsymbol{84.48}_{\pm 1.40}$ & 25.16M \\
   \Xhline{1.5pt}
  \end{tabular}
  \caption{\label{table:math14k_qwen3-8b}
  Experimental results on Qwen3-8B for mathematical reasoning benchmarks. Accuracy is used as the evaluation metric. All methods are run with three random seeds; the mean and standard deviation are reported. ``Param'' indicates trainable parameters. "CoMoL w/o CR" denotes that Core Space Routing is not applied in CoMoL.
  }
\end{table*}

\begingroup
\setlength{\tabcolsep}{4pt}
\begin{table*}[t]
\small
  \centering
  \begin{tabular}{lcccccccc}
    \Xhline{1.5pt}
    \rowcolor[HTML]{DAE0FB}
    \textbf{Models} & \textbf{AddSub} & \textbf{AQuA} & \textbf{GSM8k} & \textbf{MultiArith} & \textbf{SingleEq} & \textbf{SVAMP} & \textbf{Average} & \textbf{Param} \\
    \hline
    \hline
    \rowcolor[HTML]{D3D3D3}
    LoRA & $96.20_{\pm 0.25}$ & $41.34_{\pm 3.39}$ & $87.49_{\pm 0.86}$ & $97.83_{\pm 0.29}$ & $97.64_{\pm 0.41}$ & $90.60_{\pm 0.56}$ & $85.18_{\pm 0.86}$ & 35.39M \\ 
    MoLoRA & $96.37_{\pm 0.29}$ & $44.88_{\pm 2.73}$ & $87.59_{\pm 1.12}$ & $98.33_{\pm 0.29}$ & $97.90_{\pm 0.45}$ & $91.50_{\pm 0.69}$ & $86.10_{\pm 0.72}$ &  76.84M \\ 
    HydraLoRA & $95.78_{\pm 1.39}$ & $43.70_{\pm 2.39}$ & $88.05_{\pm 1.20}$ & $98.17_{\pm 1.04}$ & $97.77_{\pm 0.41}$ & $91.37_{\pm 0.81}$ & $85.81_{\pm 0.34}$ & 40.47M\\ 
    MoLA & $96.96_{\pm 0.67}$ & $42.26_{\pm 4.55}$ & $88.98_{\pm 0.68}$ & $98.67_{\pm 0.17}$ & $97.83_{\pm 0.52}$ & $91.63_{\pm 0.57}$ & $86.06_{\pm 0.66}$ & 153.68M \\
    AdaMoLE & $96.20_{\pm 0.25}$ & $40.55_{\pm 4.54}$ & $88.02_{\pm 1.26}$ & $98.50_{\pm 0.17}$ & $98.10_{\pm 0.23}$ & $91.13_{\pm 0.70}$ & $85.42_{\pm 0.54}$ & 78.36M \\ 
    SparseMoA &  $96.29_{\pm 0.89}$ & $44.62_{\pm 3.44}$ & $88.20_{\pm 1.02}$ & $98.11_{\pm 0.35}$ & $98.10_{\pm 0.11}$ & $91.63_{\pm 0.99}$ & $\underline{86.16}_{\pm 0.77}$ &  34.50M \\ 
    FlyLoRA & $95.27_{\pm 0.15}$ & $41.86_{\pm 2.24}$ & $88.53_{\pm 0.52}$ & $98.11_{\pm 0.35}$ & $96.72_{\pm 0.50}$ & $91.47_{\pm 0.15}$ & $85.33_{\pm 0.40}$ & 35.39M \\
    \hline
    \rowcolor[HTML]{E9F3FE}
    CoMoL w/o CR & $96.03_{\pm 0.96}$ & $45.28_{\pm 4.47}$ & $87.72_{\pm 2.46}$ & $98.11_{\pm 0.67}$ & $97.83_{\pm 0.20}$ & $91.50_{\pm 0.96}$ & $86.08_{\pm 1.23}$ & 47.90M \\
    \rowcolor[HTML]{E9F3FE}
    CoMoL & $96.62_{\pm 0.53}$ & $46.06_{\pm 4.09}$ & $87.89_{\pm 0.38}$ & $98.11_{\pm 0.35}$ & $97.44_{\pm 0.68}$ & $91.93_{\pm 0.91}$ & $\boldsymbol{86.34}_{\pm 1.01}$ & 35.82M \\
   \Xhline{1.5pt}
  \end{tabular}
  \caption{\label{table:math14k_qwen3-14b}
  Experimental results on Qwen3-14B for mathematical reasoning benchmarks. 
  }
\end{table*}

\paragraph{Datasets.}
To evaluate the performance on mathematical reasoning tasks, we utilize Math14k \cite{hu_llm-adapters_2023} as the fine-tuning corpus, which comprises the training subsets of GSM8K and AQuA, augmented with Chain-of-Thought (CoT) reasoning paths to enhance logical transparency. During the evaluation phase, we systematically benchmark our model across six diverse datasets: GSM8K \cite{cobbe2021gsm8k}, SVAMP \cite{patel2021SVAMP}, MultiArith \cite{roy2016MultiArith}, AddSub \cite{hosseini2014AddSub}, AQuA \cite{ling2017AQuA}, and SingleEq \cite{koncel2015SingleEQ}.
\paragraph{Experimental setting.}
We employ Qwen3-8B and the larger-scale Qwen3-14B \cite{yang2025qwen3} as our backbone models. To ensure a fair and rigorous comparison, all experiments, including the proposed method and baselines, are implemented within a unified framework based on the Transformers library \cite{wolf-etal-2020-transformers}, maintaining identical hyperparameter configurations wherever applicable. Following the setup in SparseMoA \cite{cao2025moa}, we apply LoRA or MoE-LoRA adaptations to the Query (Q), Key (K), Value (V), Output (O), and Down-projection modules, while excluding the Gate and Up-projection layers to optimize the memory footprint of MoE-LoRA methods. For all MoE-LoRA methods, we set the LoRA rank to 8. The number of experts is set to 8 for Qwen3-8B, and to 4 for Qwen3-14B for all methods except MoLA, in order to prevent GPU out-of-memory (OOM) issues. To maintain parity in the number of trainable parameters, the ranks for the standard LoRA baseline, DenseLoRA, and FlyLoRA are configured as 16, 256, and 32, respectively. Across all configurations, the LoRA scaling factor $\alpha$ is set to $1 \times \text{rank}$. All methods are trained for one epoch.
\paragraph{Main results.}

Tables~\ref{table:math14k_qwen3-8b} and~\ref{table:math14k_qwen3-14b} summarize the experimental results for the Qwen3-8B and Qwen3-14B backbones. Notably, CoMoL achieves a superior balance between parameter efficiency and overall performance. On the Qwen3-8B model, CoMoL reaches an average accuracy of 84.48\%, outstripping the Standard LoRA baseline by a margin of 1.7 percentage points while maintaining parameter parity. It achieves state-of-the-art (SOTA) performance on the AddSub and AQuA benchmarks, surpassing LoRA by a significant 7.2 points on the latter. Furthermore, CoMoL consistently outperforms other competitive LoRA-based variants, such as SparseMoA and FlyLoRA. Impressively, it even exceeds the performance of MoLoRA and MoLA, despite these models possessing four times the number of trainable parameters. These findings substantiate the presence of significant expert redundancy in traditional MoE-LoRA architectures and demonstrate that CoMoL can effectively capture the collective expertise of multiple LoRA experts with negligible parameter overhead.

The performance gains of CoMoL are consistently maintained across different model scales, outperforming all baselines on both the Qwen3-8B and the larger Qwen3-14B variants. Such results highlight the exceptional stability and robustness of our CoMoL across varying parameter capacities.


\begingroup
\setlength{\tabcolsep}{2pt}
\begin{table*}[t]
  \centering
  \begin{tabular}{lcccrcccr}
    \Xhline{1.5pt}
    \rowcolor[HTML]{DAE0FB}
    & \multicolumn{4}{|c}{\texttt{Llama3.1-8B}} & \multicolumn{4}{|c}{\texttt{Qwen3-8B}} \\
    \hline
    \hline
    \rowcolor[HTML]{DAE0FB}
    \textbf{Methods}  & \textbf{Pass@1} & \textbf{Pass@5} & \textbf{Pass@10} &  \textbf{\# Parameters} & \textbf{Pass@1} & \textbf{Pass@5} & \textbf{Pass@10} &  \textbf{\# Parameters} \\
    \hline
    LoRA & 26.22 & 54.41 & 67.07 & 23.06M & 39.69 & 72.29 & 81.71 & 24.77M\\
    MoLoRA & \underline{34.15} & 63.15 & \underline{72.56} & 100.14M & 43.78 & 73.49 & 82.93 & 107.35M\\
    HydraLora & 33.35 & 60.96 & 69.51 & 45.09M & \underline{46.89} & 75.61 & 84.15 & 49.55M\\
    MoLA & 28.72 & 56.53 & 67.68 & 100.14M & 41.10 & 71.60 & 81.10 & 107.35M \\
    AdaMoLE & 21.52 & 44.58 & 56.10 & 101.12M & 40.91 & 66.65& 76.83 & 108.38M \\
    DenseLoRA & 31.58 & 58.71 & 67.07 & 26.74M & 46.82 & 76.04 & 84.75 & 27.00M \\
    FlyLoRA & 32.32 & \underline{63.21}	& 71.95 & 23.06M & 20.18 & 38.15 & 48.17 & 24.77M \\
    \hline
    \rowcolor[HTML]{E9F3FE}
    CoMoL w/o CR & 31.10 & 59.42 & 68.90 & 27.16M & 43.23 & \textbf{77.25} & \textbf{87.19} & 33.40M \\ 
    \rowcolor[HTML]{E9F3FE}
    CoMoL & \textbf{35.00} & \textbf{63.65} & \textbf{73.17} & 23.24M & \textbf{48.11} & \underline{77.22} & \underline{86.59} & 24.97M \\ 
    \Xhline{1.5pt}
  \end{tabular}
  \caption{\label{table:code_result}
    Experimental results for the code generation task on Llama3.1-8B and Qwen3-8B. Pass@k is used as the evaluation metric. "\# Parameters" indicates trainable parameters. Bold numbers indicate the highest performance scores, and underlined numbers indicate the second-highest performance scores.
  }
\end{table*}
\endgroup

\subsection{Code Generation Task}
\paragraph{Datasets.}
To evaluate the code generation proficiency of our method, we adopt the experimental protocol established by FlyLoRA \cite{zouflylora} and utilize the CodeAlpaca-20k \cite{chaudhary2023codealpaca} dataset as our instruction-tuning corpus. The evaluation is conducted on the HumanEval \cite{chen2021humaneval} benchmark. To provide a holistic and comprehensive assessment of the model's coding capabilities, we report the pass@$k$ metrics (where $k \in \{1, 5, 10\}$), which measure the probability of generating at least one functionally correct solution within $k$ attempts.
\paragraph{Experimental setting.}
For the code generation task, we employ Qwen3-8B and Llama3.1-8B as our backbone models, adopting hyperparameter configurations analogous to those used in the mathematical reasoning experiments. Recognizing that various baselines exhibit high sensitivity to training duration in coding tasks, we perform model selection by searching for the optimal checkpoint based on validation set performance. Specifically, the models are trained for 5 epochs on Qwen3-8B and 1 epoch on Llama3.1-8B, ensuring that each method is evaluated at its peak proficiency.
\paragraph{Main results.}



CoMoL achieves or surpasses other methods with minimal trainable parameters, showing its parameter efficiency. As shown in the Table~\ref{table:code_result}, methods with more trainable parameters, MoLA and AdaMoLE, actually perform worse than HydraLoRA and DenseLoRA with fewer parameters, indicating that methods with excessive parameters are prone to overfitting on the code generation task. In contrast, CoMoL maintains strong performance on both base models.

We observe that FlyLoRA's sparse learning approach with rank-wise expert activation in the up-projection matrix of LoRA performs well on Llama3.1-8B but poorly on Qwen3-8B. This is because FlyLoRA's sparsity reduces its learning capacity. The code generation training and test sets are both instruction fine-tuning datasets, which align closely with Llama3.1-8B's original capability distribution. However, Qwen3-8B is a long-thinking model, requiring the adaptation module to have sufficient capacity to modify its long-thinking behavior. Although CoMoL uses only a small number of expert parameters in the core space, its learning capacity remains stable, surpassing other MoE-LoRA methods and DenseLoRA.

\subsection{Scaling with Rank}
Figure~\ref{fig:comol_rank} presents the performance and trainable parameter count of CoMoL and Standard LoRA across different ranks on the code generation task. \textbf{CoMoL consistently outperforms LoRA across all ranks, achieving the best performance at rank 16}. CoMoL w/o CR surpasses LoRA at all ranks except rank 32, and even exceeds CoMoL at rank 8. Notably, with a fixed number of experts, CoMoL maintains a parameter count comparable to LoRA while achieving consistent improvements.

\subsection{Scaling with Expert Number}

Table~\ref{table:experts_scale} reports the performance of CoMoL and CoMoL w/o CR with varying numbers of experts on the mathematical reasoning benchmark. On both Qwen3-8B and Qwen3-14B, the two methods achieve their best performance with 8 experts. Notably, both methods scale efficiently to 64 experts, whereas other approaches, such as HydraLoRA, encounter out-of-memory (OOM) issues with only 16 experts under the same GPU memory constraints.

\begin{figure*}[ht]
  \includegraphics[width=1\textwidth]{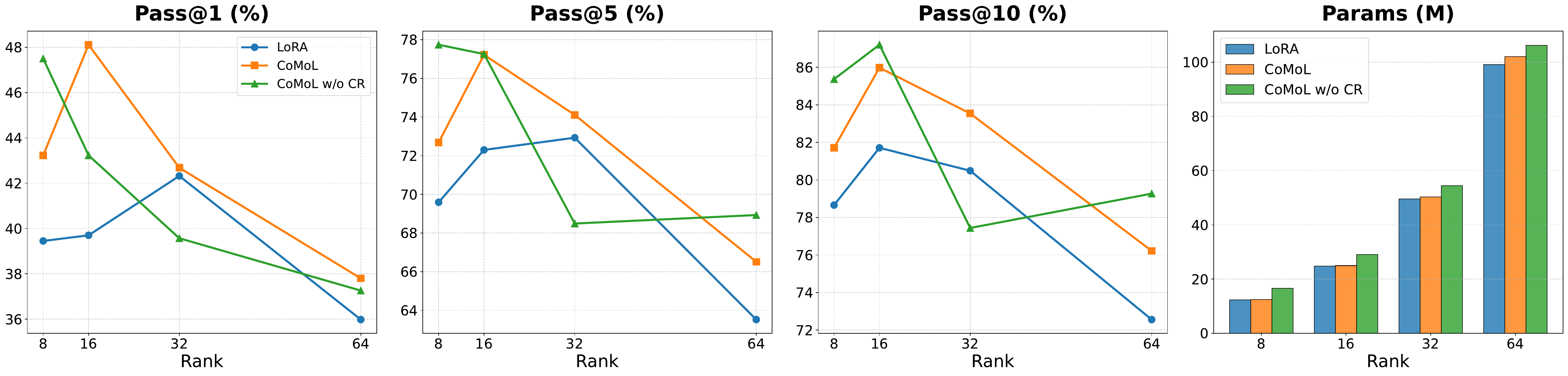}
  \caption{
    \textbf{
    Performance and trainable parameter comparison between CoMoL and LoRA across different ranks on the code generation benchmark.}
    }
  \label{fig:comol_rank}
\end{figure*}

\textbf{Core Space Routing reduces router parameters while maintaining stable performance}. Comparing the accuracy and trainable parameter count of CoMoL and CoMoL w/o CR across different numbers of experts, we observe that the two methods achieve comparable performance. However, CoMoL with Core Space Routing introduces virtually no additional router parameters as the number of experts increases. This demonstrates that Core Space Routing effectively leverages the low-rank property of LoRA.

\begingroup
\begin{table}[t]
  \centering
  \begin{tabular}{lcrcr}
    \Xhline{1.5pt}
    \rowcolor[HTML]{DAE0FB}
    & \multicolumn{2}{|c}{\texttt{CoMoL w/o CR}} & \multicolumn{2}{|c}{\texttt{CoMoL}} \\
    \hline
    \hline
    \rowcolor[HTML]{DAE0FB}
    \textbf{\# Experts}  & \textbf{Math}  &  \textbf{\# Param} & \textbf{Math} &  \textbf{\# Param} \\
    \hline
    \hline
    \rowcolor[HTML]{E9F3FE}
    \multicolumn{5}{c}{\texttt{Qwen3-8B}} \\
    \hline
    8 & 84.47 & 33.39M & 84.48 & 25.16M \\
    16 & 82.50 & 42.02M & 84.00 & 25.56M\\
    32 & 82.49 & 59.28M & 82.24 & 26.34M \\
    64 & 84.50 & 93.78M & 82.72 & 27.91M \\
    \hline
    \hline
    \rowcolor[HTML]{E9F3FE}
    \multicolumn{5}{c}{\texttt{Qwen3-14B}} \\
    \hline
    8 & 86.08 & 47.90M & 86.34 & 35.82M \\
    16 & 85.92 & 60.46M & 86.24 & 36.26M \\
    32 & 86.13 & 85.52M & 86.06 & 37.13M \\
    64 & 86.32 & 135.66M & 86.23 & 38.87M \\
    \Xhline{1.5pt}
  \end{tabular}
  \caption{\label{table:experts_scale}
    Comparison between CoMoL and CoMoL w/o CR with varying numbers of experts on mathematical reasoning. "\# Param" indicates trainable parameters. 
  }
\end{table}
\endgroup

Notably, scaling the number of experts to 64 does not yield consistent improvements over 8 experts. This observation mirrors a well-known phenomenon in Standard LoRA, where simply increasing the rank does not necessarily translate to proportional performance gains \cite{chen2022revisiting}. We attribute this to diminishing returns in expert scaling under data-constrained settings: when the quality and quantity of training samples are limited, a larger pool of experts cannot be adequately optimized, resulting in under-training and degraded specialization.

\subsection{Computational Efficiency Analysis}

\begingroup
\setlength{\tabcolsep}{2pt}
\begin{table}[t]
\centering
\begin{tabular}{lcc}
\Xhline{1.5pt}
\rowcolor[HTML]{DAE0FB}
\textbf{Method} & \textbf{Training Time} & \textbf{Inference Time}\\
\hline
\hline
\rowcolor[HTML]{D3D3D3}
LoRA         & 00:56:47 & 18.40 \\
MoLoRA       & 04:44:50 & 19.67 \\
HydraLoRA    & 04:15:15 & 19.16 \\
MoLA         & 07:59:02 & 30.03 \\
AdaMoLE      & 08:18:54 & 40.78 \\
\hline
\rowcolor[HTML]{E9F3FE}
CoMoL w/o CR & 02:33:27 & 18.62 \\
\rowcolor[HTML]{E9F3FE}
CoMoL        & 02:19:55 & 18.57 \\
\Xhline{1.5pt}
\end{tabular}
\caption{
Wall-clock time comparison of training and inference across methods. Training time (HH:MM:SS) is measured for one epoch on CodeAlpaca-20k (batch size 1). Inference time (minutes) is the cumulative duration for 10 generations per sample on HumanEval (batch size 16). All MoE-LoRA variants use $k{=}8$ experts.
} \label{tab:efficiency}
\end{table}
\endgroup

Table~\ref{tab:efficiency} reports the wall-clock training and inference times of CoMoL alongside representative baselines under controlled conditions. To ensure a fair comparison, we measure training time as the total duration to complete one epoch on the CodeAlpaca-20k dataset
with a batch size of 1, and inference time as the cumulative duration for 10 generations per sample on the HumanEval benchmark. Following the experimental protocol described in our main experiments, all MoE-based variants employ $k{=}8$ experts. 

As shown in the table, CoMoL achieves substantially lower training cost than all competing MoE-LoRA methods, roughly half that of MoLoRA and HydraLoRA, while maintaining inference latency comparable to standard LoRA and consistently below that of other MoE-LoRA counterparts. The advantage is particularly pronounced against the two sparse MoE-LoRA methods (MoLA and AdaMoLE), whose training
times exceed CoMoL's by approximately $4\times$ and whose inference times are $1.5\times$ to $2\times$ longer. These empirical wall-clock measurements corroborate our theoretical FLOPs analysis presented in Section~\ref{sec:flops}, confirming that the efficiency gains of CoMoL translate directly into practical speedups.



\section{Related work}

\paragraph{Sparse MoE-LoRA.}
Sparse MoE-LoRA has been studied from multiple angles, including routing regularization, expert diversity, and redundancy reduction. SiRA \citep{zhu_sira_2023} adopts top-$k$ routing with per-expert capacity constraints and gating dropout to stabilize routing and alleviate overfitting. MoELoRA \cite{luo_moelora_2024} encourages expert specialization via contrastive learning to mitigate random routing. MoLA \citep{gao_higher_2024} analyzes layer-wise redundancy by allocating different numbers of LoRA experts across layers, showing higher redundancy in lower layers. AdaMoLE \citep{liu_adamole_2024} further introduces token-wise adaptive expert activation via a threshold network integrated into top-$k$ routing.

\paragraph{Soft-weighted MoE-LoRA.}
These variants replace hard top-$k$ selection with continuous mixtures over LoRA experts. MoLoRA \cite{zadouri_pushing_2023} adopts token-level soft routing to combine experts per token. LoRAMoE \cite{dou_loramoe_2024} uses a linear router to integrate LoRA experts and mitigate forgetting of world knowledge. MOLE \cite{wu_mixture_2024} enables efficient composition of multiple trained LoRAs by training only the per-layer router. HydraLoRA \cite{tian_hydralora_2024} further improves parameter efficiency via an asymmetric design with a shared matrix.

 \paragraph{Soft-merging MoE methods.}
In contrast to methods that aggregate expert outputs, some approaches (AdaMix, SMEAR, MoV) adopt an alternative strategy of first merging expert parameters via routing before performing computation, thereby reducing computational overhead. For instance, AdaMix \cite{wang_adamix_2022} employs a mixture of adapters with stochastic routing, while SMEAR \cite{muqeeth_soft_2024} proposes merging experts through a learned router. Notably, both AdaMix and SMEAR operate at the instance level. MoV \cite{zadouri_pushing_2023}, on the other hand, is a token-level expert parameter merging method. However, it is limited to IA3 \cite{liu2022IA3} experts, which scale hidden states in the base model using trainable vectors. Currently, no existing method is capable of merging LoRA experts at the token level.
\section{Conclusion}


In this work, we address the limited parameter efficiency and coarse-grained adaptation of existing MoE-LoRA architectures by proposing Core Space Mixture of LoRA (CoMoL), a novel framework built upon two key components: Core Space Experts and Core Space Routing. By operating entirely within a compact core space, CoMoL enables fine-grained, input-adaptive expert merging while maintaining parameter and computational costs comparable to Standard LoRA. Moreover, projecting the routing network into the same low-rank space further reduces overhead without compromising expressiveness. Extensive experiments demonstrate that CoMoL consistently outperforms existing MoE-LoRA methods across diverse tasks, highlighting its effectiveness, scalability, and practical value for parameter-efficient adaptation of LLMs.


\section*{Limitations}
Current research on MoE-LoRA and other PEFT methods primarily focuses on improving parameter efficiency, while the learning capacity of these methods across different fine-tuning scenarios remains underexplored. For instance, FlyLoRA exhibits substantially different performance on the code generation task when applied to Llama3.1-8B versus Qwen3-8B. There currently exists no systematic benchmark or evaluation framework to assess the learning capacity of different PEFT methods. Although CoMoL demonstrates strong performance across various tasks (e.g., mathematical reasoning and code generation), different model families, and varying model scales, a more comprehensive understanding of the capacity boundaries of different PEFT methods remains to be established. We leave this systematic investigation as future work.

\section*{Acknowledgments}
This work was supported by the National Key R\&D Program of China (2025ZD0123100), the NSFC (62272411), the Key R\&D Projects in Zhejiang Province (No. 2025C01030, No. 2025C01128), the Zhejiang NSF (LRG25F020001), the Postdoctoral Fellowship Program of CPSF (GZC20251077), the China Postdoctoral Science Foundation (2025M781525), Fundamental Research Funds for the Central Universities (226-2025-00057), and the Tencent WeChat Rhino-Bird Special Research Program (Tencent WXG-FR-2023-10).

\bibliography{custom}

\appendix


\end{document}